%% file: egpaper_for_review.tex
\documentclass[10pt,twocolumn,letterpaper]{article}

\usepackage{iccv}
\usepackage{times}
\usepackage{epsfig}
\usepackage{graphicx}
\usepackage{amsmath}
\usepackage{amssymb}

\usepackage[export]{adjustbox}
\usepackage[dvipsnames]{xcolor}
\usepackage{multirow}

\definecolor{purple}{RGB}{230,0,250}

\usepackage{environ}
\newcommand{\acksection}{\section*{Acknowledgments}}
\NewEnviron{acks}{%
  \acksection
  \BODY
}

\usepackage[pagebackref=true,breaklinks=true,letterpaper=true,colorlinks,bookmarks=false]{hyperref}

\iccvfinalcopy % *** Uncomment this line for the final submission

\def\iccvPaperID{} % *** Enter the ICCV Paper ID here

\ificcvfinal\fi

\begin{document}

%%%%%%%%% TITLE
\title{Adaptive Surface Normal Constraint for Depth Estimation}

\author{Xiaoxiao Long$^{1}$ \quad Cheng Lin$^{1}$ \quad Lingjie Liu$^{2}$ \quad Wei Li$^{3}$ \\
Christian Theobalt$^{2}$ \quad Ruigang Yang$^{3}$ \quad Wenping Wang$^{1,4}$ \\[0.3em]
$^{1}$The University of Hong Kong \quad $^{2}$Max Planck Institute for Informatics \\
$^{3}$Inceptio \quad $^{4}$Texas A\&M University }

\maketitle
% Remove page # from the first page of camera-ready.
\ificcvfinal\thispagestyle{empty}\fi

\input{0_abs}

\input{1_intro}

\input{2_related_works}

\input{3_method}

\input{4_implementation}

\input{5_0_experiments}

\input{5_1_discussions}

\input{6_conclusion}

{\small
\bibliographystyle{ieee_fullname}
\bibliography{egbib}

\input{egbib.bbl}
}

\end{document}

%% file: 0_abs.tex
%%%%%%%%% ABSTRACT
\begin{abstract}
We present a novel method for single image depth estimation using surface normal constraints. Existing depth estimation methods either suffer from the lack of geometric constraints, or are limited to the difficulty of reliably capturing geometric context, which leads to a bottleneck of depth estimation quality. We therefore introduce a simple yet effective method, named Adaptive Surface Normal (ASN) constraint, to effectively correlate the depth estimation with geometric consistency. Our key idea is to adaptively determine the reliable local geometry from a set of randomly sampled candidates to derive surface normal constraint, for which we measure the consistency of the geometric contextual features. As a result, our method can faithfully reconstruct the 3D geometry and is robust to local shape variations, such as boundaries, sharp corners and noises. We conduct extensive evaluations and comparisons using public datasets. The experimental results demonstrate our method outperforms the state-of-the-art methods and has superior efficiency and robustness~\footnote{Codes will be available at: \url{https://github.com/xxlong0/ASNDepth}}.

%We present a novel method for single image depth estimation, which plays a fundamental role in various applications, such as scene understanding, robot navigation and novel view synthesis. Although recent methods have achieved high accuracy in pixel-wise error metrics, their depth maps are often distorted and fail to keep global and local geometric characteristics in 3D space. Our method achieves geometric preserving results by introducing a Pixel-Adaptive Surface Normal constraint (PASN), which could enforce the constraint by considering the content of the input image. Guided by the learnable pixel features of the input image, PASN could perform boundary-aware surface normal calculation and avoid inaccurate supervision near edges. Moreover, compared with least square based surface normal calculation operations, PASN is faster and computational efficient.

\end{abstract}

%% file: 1_intro.tex
\section{Introduction}

\begin{figure}[htp]
\setlength{\abovecaptionskip}{0pt}
\setlength{\belowcaptionskip}{0pt}
    \centering
    \includegraphics[width=\columnwidth]{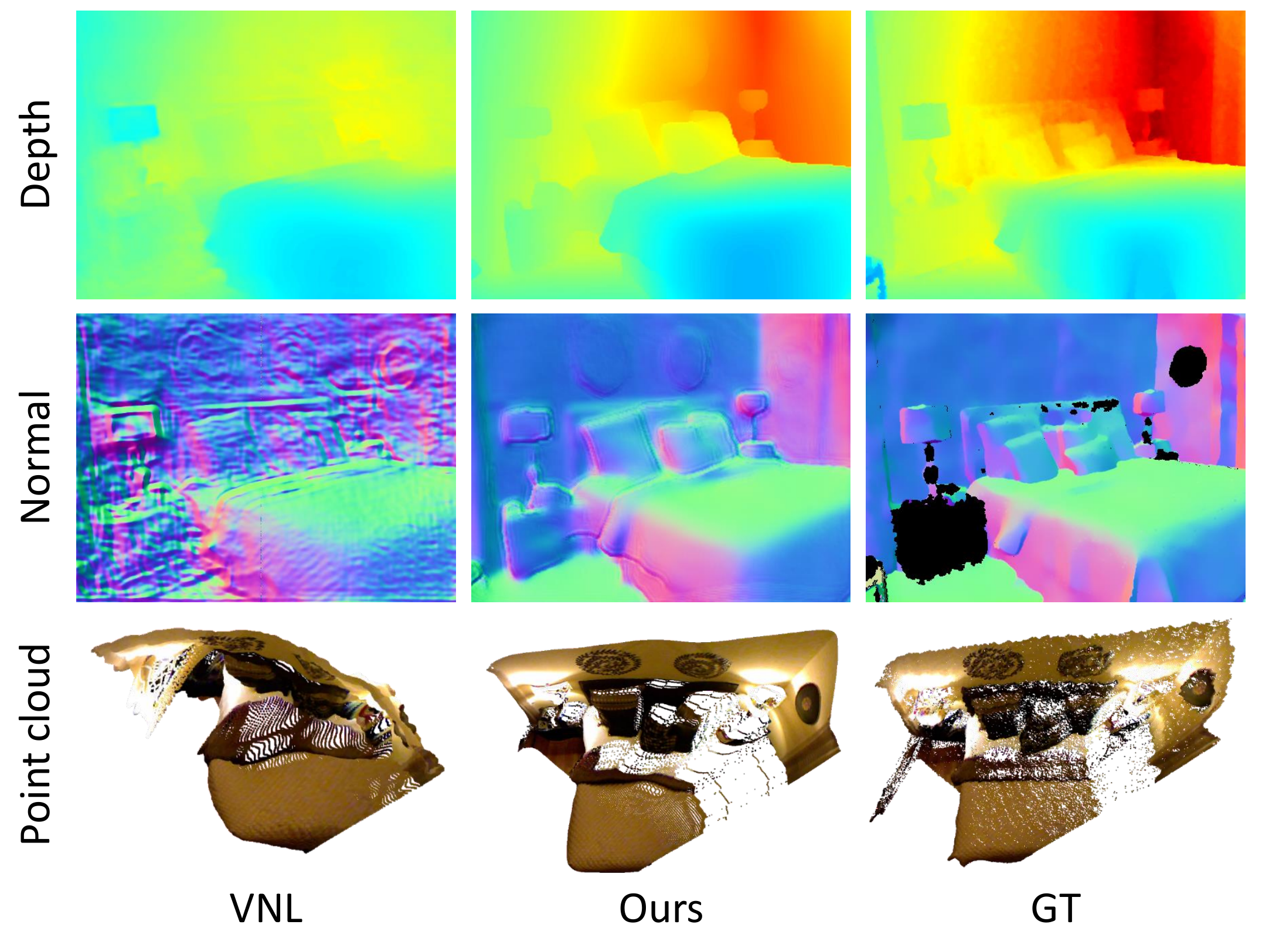}
    \caption{Example results of ground truth, ours and VNL~\cite{yin2019enforcing}. By enforcing our proposed Adaptive Surface Normal (ASN) constraint, our reconstructed point cloud preserves both global structural information and local geometric features. The recovered surface normal is more accurate and less noisy than that of VNL.}
    \label{fig:teaser}
\end{figure}

Estimating depth from a single RGB image, one of the most fundamental computer vision tasks, has been extensively researched for decades. With the recent advances of deep learning, depth estimation using neural networks has drawn increasing attention. Earlier works \cite{eigen2014depth,liu2015learning,fu2018deep,roy2016monocular,ranftl2016dense} in this field directly minimize the pixel-wise depth errors, of which results cannot faithfully capture the 3D geometric features. Therefore, the latest efforts incorporate geometric constraints into the network and show promising results. 

Among various geometric attributes, surface normal is predominantly adopted due to the following two reasons. First, surface normal can be estimated by the 3D points converted from depth. Second, surface normal is determined by a surface tangent plane, which inherently encodes the local geometric context. Consequently, to extract normals from depth maps as geometric constraints, previous works propose various strategies, including random sampling~\cite{yin2019enforcing}, Sobel-like operator~\cite{hu2019revisiting, kusupati2020normal} and differentiable least square~\cite{qi2018geonet, long2020occlusion}. 

Despite the improvements brought about by the existing efforts, a critical issue remains unsolved, i.e., how to determine the reliable local geometry to correlate the normal constraint with the depth estimation. For example, at shape boundaries or corners, the neighboring pixels for a point could belong to different geometries, where the local plane assumption is not satisfied. Due to this challenge, these methods either struggle to capture the local features~\cite{yin2019enforcing}, or sensitive to local geometric variations (noises or boundaries)~\cite{hu2019revisiting, kusupati2020normal}, or computationally expensive~\cite{qi2018geonet, long2020occlusion}.

Given the significance of local context constraints, there is a multitude of works on how to incorporate shape-ware regularization in monocular reconstruction tasks, ranging from sophisticated variational approaches for optical flow~\cite{revaud2015epicflow,xiao2006bilateral,ren2008local,bao2014fast} to edge-aware filtering in stereo~\cite{song2018edgestereo} and monocular reconstruction~\cite{jiao2014local,qi2020geonet++}. However, these methods have complex formulations and only focus on 2D feature edges derived from image intensity variation, without considering geometric structures of shapes in 3D space.

In this paper, we introduce a simple yet effective method to correlate depth estimation with surface normal constraint. Our formulation is much simpler than any of the aforementioned approaches, but significantly improves the depth prediction quality, as shown in Fig.~\ref{fig:teaser}. Our key idea is to adaptively determine the faithful local geometry from a set of randomly sampled candidates to support the normal estimation. For a target point on the image, first, we randomly sample a set of point triplets in its neighborhood to define the candidates of normals. Then, we determine the confidence score of each normal candidate by measuring the consistency of the learned latent geometric features between the candidate and the target point. Finally, the normal is adaptively estimated as a weighted sum of all the candidates. 

Our simple strategy has some unique advantages: 1) the random sampling captures sufficient information from the neighborhood of the target point, which is not only highly efficient for computation, but also accommodates various geometric context; 2) the confidence scores adapatively determine the reliable candidates, making the normal estimation robust to local variations, e.g., noises, boundaries and sharp changes; 3) we measure the confidence using the learned contextual features, of which representational capacity is applicable to complex structures and informative to correlate the normal constraint with the estimated depth. More importantly, our method achieves superior results on the public datasets and considerably outperforms the state-of-the-art methods.

Our main contributions are summarized as follows:

\begin{itemize}
    \vspace{-2mm}
    \item We introduce a novel formulation to derive geometric constraint for depth estimation, i.e., adaptive surface normal.
    \vspace{-2mm}
    \item Our method is simple, fast and effective. It is robust to noises and local variations and able to consistently capture faithful geometry.
    \vspace{-2mm}
    \item Our method outperforms the state-of-the-art method on the public datasets by a large margin.
\end{itemize}

%% file: 2_related_works.tex
\begin{figure*}[!t]
\setlength{\abovecaptionskip}{0pt}
\setlength{\belowcaptionskip}{0pt}
    \centering
    \includegraphics[width=\linewidth]{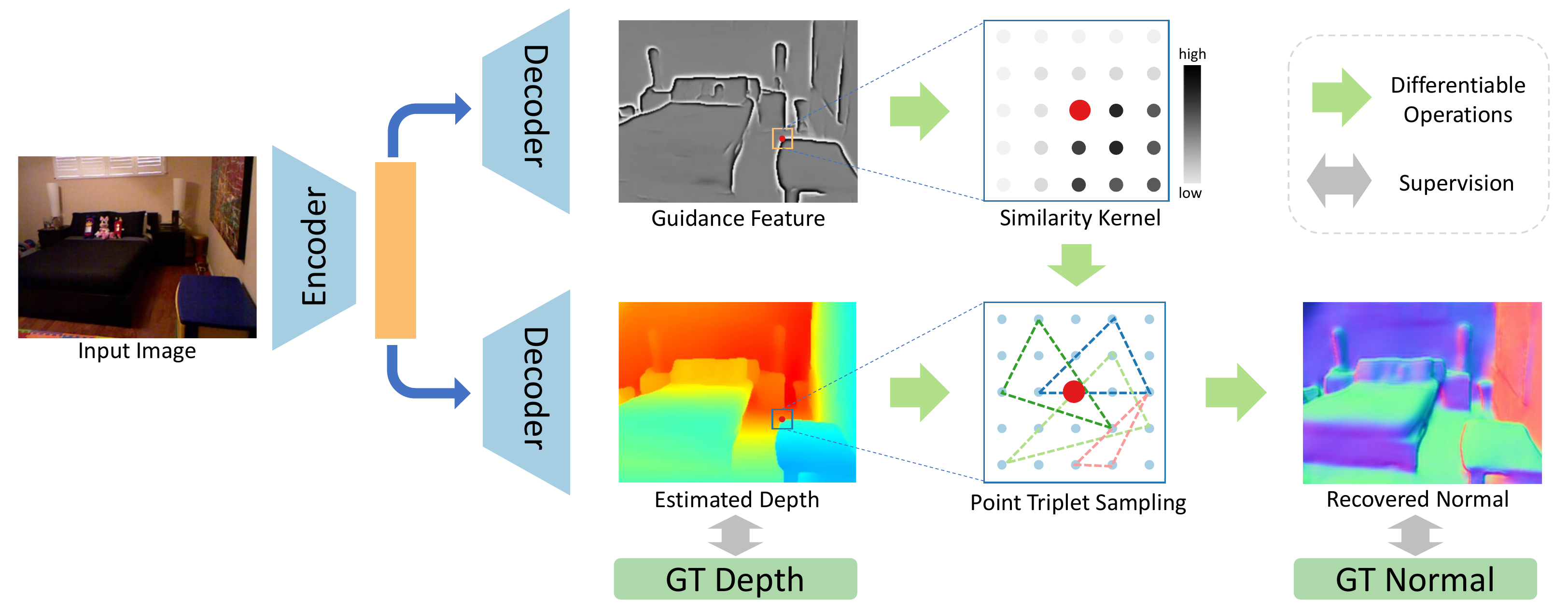}
    \caption{Overview of our method. Taking a single image as input, our model produces estimated depth and guidance feature from two decoders, respectively. We recover surface normal from the estimated depth map with our proposed Adaptive Surface Normal (ASN) computation method. The similarity kernels computed from guidance feature enable our surface normal calculation to be local geometry aware, like shape boundaries and corners. Finally, pixel-wise depth supervision is enforced on the estimated depth, while the geometric supervision is enforced on the recovered surface normal.}
    \label{fig:pipeline}
    \vspace{-3mm}
\end{figure*}

\vspace{-3mm}
\section{Related Work}
\paragraph{Monocular depth estimation} As an ill-posed problem, monocular depth estimation is challenging, given that minimal geometric information can be extracted from a single image. Recently, benefiting from the prior structural information learned by the neural network, many learning-based works~\cite{eigen2014depth,liu2015learning,xu2017multi,fu2018deep,roy2016monocular,ranftl2016dense,godard2019digging,godard2017unsupervised,liu2020fcfr,liu2021learning} have achieved promising results.
Eigen~\etal~\cite{eigen2014depth} directly estimate depth maps by feeding images into a multi-scale neural network. Laina~\etal~\cite{laina2016deeper} propose a deeper residual network and further improve the accuracy of depth estimation. Liu~\etal~\cite{liu2015learning} utilize a continuous conditional random field (CRF) to smooth super-pixel depth estimation. Xu~\etal~\cite{xu2017multi} propose a sequential network based on multi-scale CRFs to estimate depth. Fu~\etal~\cite{fu2018deep} design a novel ordinal loss function to recover the ordinal information from a single image. Unfortunately, the estimated depth maps of these methods always fail to recover important 3D geometric features when converted to point clouds, since these methods do not consider any geometric constraints.

\vspace{-3mm}
\paragraph{Joint depth and normal estimation} Since the depth and surface normal are closely related in terms of 3D geometry, there has been growing interests in joint depth and normal estimation using neural networks to improve the performance. Several works \cite{eigen2015predicting,zhang2019pattern,xu2018pad,li2015depth} jointly estimate depth and surface normal using multiple branches and propagate the latent features of each other. Nevertheless, since there are no explicit geometric constraints enforced on the depth estimation, the predicted geometry of these methods is still barely satisfactory. 

Consequently, methods~\cite{xu2019depth,yang2018unsupervised,qiu2019deeplidar,hu2019revisiting,qi2018geonet,long2020occlusion,kusupati2020normal} are proposed to explicitly enforce geometric constraints on estimated depth maps.. Hu~\etal~\cite{hu2019revisiting} and Kusupati~etal~\cite{kusupati2020normal} utilize a Sobel-like operator to calculate surface normals from estimated depths, and then enforce them to be consistent with the ground truth. Nonetheless, the Sobel-like operator can be considered as a fixed filter kernel that indiscriminately acts on the whole image (see Fig.~\ref{fig:normal_diagram}), leading to unacceptable inaccuracy and sensitivity to noises. To constrain surface normal more reliably, Qi~\etal~\cite{qi2018geonet} and Long~\etal~\cite{long2020occlusion} propose to utilize a differentiable least square module for surface normal estimation. These methods optimize the geometric consistency, of which solution is more accurate and robust to noises but limited to expensive computation. Yin~\etal~\cite{yin2019enforcing} introduce virtual normal, which is a global geometric constraint derived from the randomly sampled point triplets from estimated depth. However, this constraint struggles to capture local geometric features, given that the point triplets are randomly sampled from the whole image.

\paragraph{Edge preserving methods} 
Out of the statistical relations between shape boundaries and image intensity edges, many works leverage this statistic prior to benefit many vision tasks.
Anisotropic diffusion~\cite{perona1990scale,black1998robust,weickert1998anisotropic} is a well-known technique for image denoising without removing important details of image content, typically edges.
Works~\cite{revaud2015epicflow,xiao2006bilateral,ren2008local,bao2014fast} propose variational approaches with anisotropic diffusion to model local edge structures for optical flow estimation. Some stereo/monocular depth estimation works rely on pretrained edge detection network~\cite{song2018edgestereo} or Canny edge detector~\cite{jiao2014local,qi2020geonet++}, to extract image edges to improve depth estimation. 
However, only a small fraction of the intensity edges keep consistent with true geometric shape boundaries. Our method could detect the true shape boundaries where 3D geometry changes instead of image intensity edges.

%% file: 3_method.tex
\section{Method}

Given a single color image $I$ as input, we use an encoder-decoder neural network to output its depth map $D_{pred}$. Our approach aims to not only estimate accurate depth but also recover high-quality 3D geometry. For this purpose, we correlate surface normal constraint with depth estimating. Overall, we enforce two types of supervision for training the network. First, like most of depth estimation works, we employ a pixel-wise depth supervision like $L_1$ loss over the predicted depth $D_{pred}$ and ground truth depth $D_{gt}$. Moreover, we compute the surface normal $N_{pred}$ from $D_{pred}$  using an \textit{adaptive strategy}, and enforce consistency between $N_{pred}$ with the ground truth surface normal $N_{gt}$, named as Adaptive Surface Normal (ASN) constraint. The method is overviewed in Fig.~\ref{fig:pipeline}.

% \vspace{-4mm}
\paragraph{Local plane assumption.} To correlate surface normal constraint with depth estimation, we adopt the local plane assumption following ~\cite{qi2018geonet,long2020occlusion}. That is, a small set of neighborhoods of a point forms a local plane, of which normal vector approximates the surface normal. Hence, for a pixel on the depth map, its surface normal can be estimated by the local patch formed by its neighboring points. In theory, the local patch could have arbitrary shapes and sizes. In practice, however, square local patches are widely adopted with sizes $(2m+1)\times(2m+1), m=1,2,...,n$, due to its simplicity and efficiency.

%a local patch around the target pixel will be selected, whose pixels are sampled to estimate the surface normal.
% \vspace{-4mm}
\paragraph{Normal candidates sampling.} To compute the surface normal, unlike prior works utilize least square fitting ~\cite{qi2018geonet,long2020occlusion} or Sobel-like kernel approximation ~\cite{hu2019revisiting,kusupati2020normal}, we propose a randomly sampling based strategy. 

For a target point $P_i\in\mathbb{R}^3$, we first extract all the points $\mathbb{P}_i=\left\{P_{j} \mid P_j\in\mathbb{R}^3,\ j=0,\ldots, r^2-1\right\}$ within a local patch of size $r \times r$. Then, we randomly sample $K$ point triplets in $\mathbb{P}_i$. All sampled point triplets for the target point $P_i$ are denoted as $\mathbb{T}_i=\left\{\left(P_k^A, P_k^B, P_k^C\right) \mid P\in \mathbb{R}^3,\ k=0,\ldots ,K-1\right\}$.
% and their corresponding 2D pixel coordinates in the depth map are $\left\{\left(p_{a}, p_{b}, p_{c}\right)_{i} \mid p\in\mathbb{R}^2,\ i=0,\ldots,N-1\right\}$. 
If the three points are not colinear, the normal vector of the sampled local plane can be directly computed by the cross-product:
\begin{equation}
\label{cross_normal}
\vec{n}_{k}=\frac{\overrightarrow{P_k^A P_k^B} \times \overrightarrow{P_k^A P_k^C}}{\mid \overrightarrow{P_k^A P_k^B} \times \overrightarrow{P_k^A P_k^C} \mid}.
\end{equation}

A normal vector will be flipped according to the viewing direction if it does not match the camera orientation. In this way, for each target point, we obtain $K$ normal candidates corresponding to $K$ sampled local planes. Next, we adaptively determine the confidence of each candidate to derive the final normal estimation result.

\begin{figure}[t]
\setlength{\abovecaptionskip}{0pt}
\setlength{\belowcaptionskip}{0pt}
    \centering
    \includegraphics[width=0.9\linewidth]{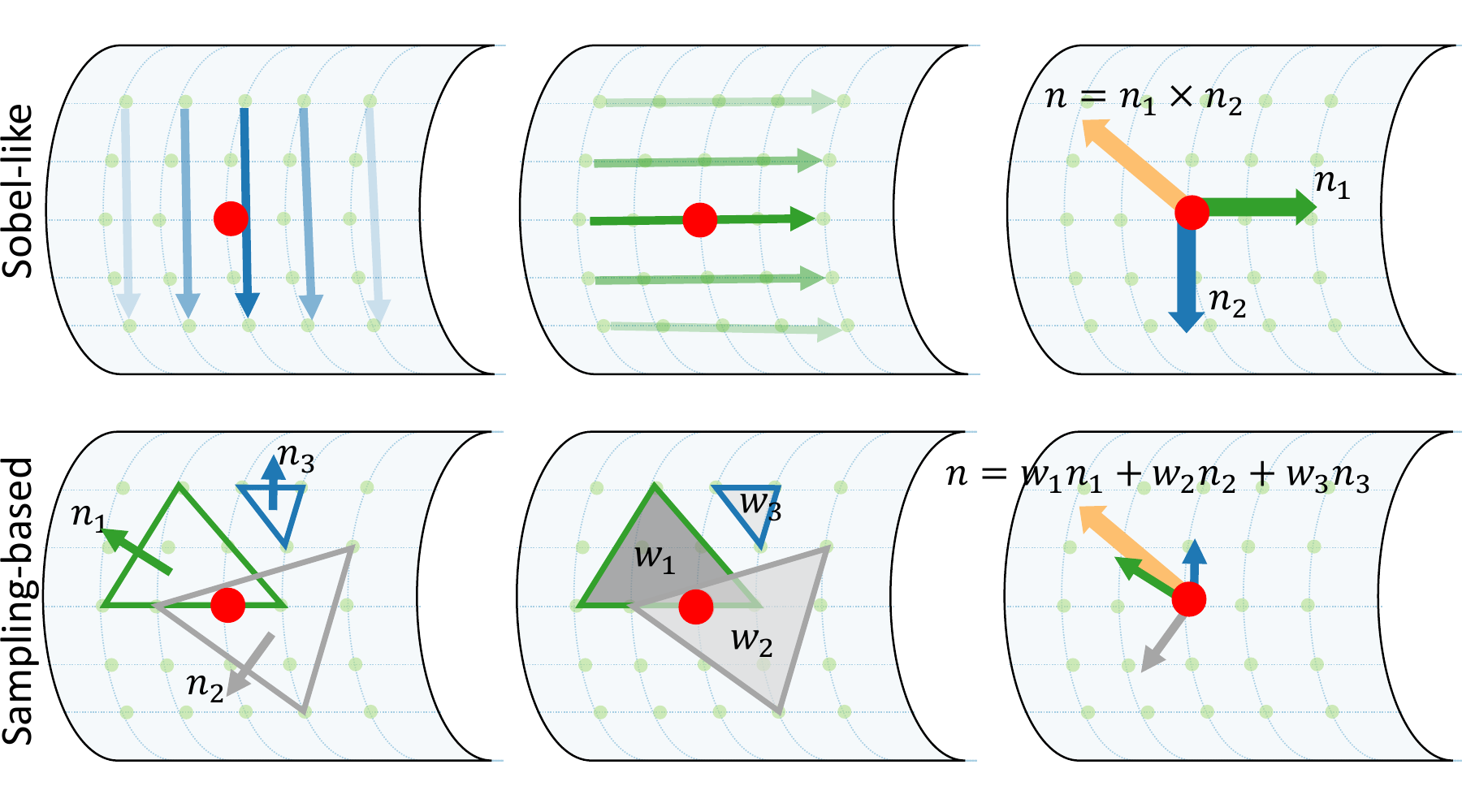}
    \caption{Sobel-like operator versus ours for surface normal estimation. The Sobel-like operator first calculates two principle vectors along up-down and left-right directions, and then use their cross product to estimate the normal. Ours first computes the normal vectors of the randomly sampled triplets, and then adaptively combines them together to obtain the final estimation.}
    \label{fig:normal_diagram}
    \vspace{-2mm}
\end{figure}

\vspace{-4mm}
\paragraph{Geometric context adaption.} 
We observe that the neighbors of a target point may not lie in the same tangent plane, especially at a region where the geometry changes, e.g., shape boundaries or sharp corners. Thus, we propose to learn a guidance feature map that is context aware to reflect the geometric variation. Therefore, the network can determine the confidence of the neighboring geometry by measuring the learned context features.

% Let $p_{t}$ represents the 2D pixel location of target point $P_{t}$, and $p_{q}$ the 2D pixel location of sampled point $P_{q}$ within the local patch of $P_{t}$, 
Given the learned guidance feature map, we measure the $L_2$ distance of the features of a sampled point $P_{j}$ and the target point $P_{i}$, and then use a normalized Gaussian kernel function to encode their latent distance into $[0,1]$:
\begin{equation}
\label{kernel_function}
\begin{aligned}
\mathcal{L}\left(P_{i}, P_{j}\right) &= e^{-0.5\left\|f\left(P_{i}\right)-f\left(P_{j}\right)\right\|_{2}} \\
\overline{\mathcal{L}}\left(P_{i}, P_{j}\right) &= \frac{\mathcal{L}\left(P_{i}, P_{j}\right)}{\sum_{P_n \in \mathbb{P}_i} \mathcal{L}\left(P_{i}, P_{n}\right)},
\end{aligned}
\end{equation}
where $f(\cdot)$ is the guidance feature map, $\left\| \cdot \right\|_{2}$ is $L_2$ distance, and $\mathbb{P}_i$ is the neighboring point set in the local patch of $P_i$ as aforementioned. E.q.~\ref{kernel_function} gives a confidence score, where the higher the confidence is, the more likely the two is to locate in the same tangent plane with target pixel. Accordingly,  the confidence score of a local plane $(P_k^A, P_k^B, P_k^C)$ to the center point $P_i$ given by the geometric adaption is defined by:

\begin{equation}
\label{geo_adaption}
g_k = \prod_{t=A,B,C} \overline{\mathcal{L}}\left(P_{i}, P_{k}^t\right).
\end{equation}

This is the multiplication of three independent probabilistic scores of the three sampled points, which measures the reliability of a sampled local plane.

\vspace{-4mm}
\paragraph{Area adaption.} The area of a sampled local plane (triangle) is an important reference to determine the reliability of the candidate. A larger triangle captures more information and thus would be more robust to local noise, as shown in ~\cite{yin2019enforcing}. For a triangle $T_k$, we simply consider its projected area $s_k$ on the image as a measurement of confidence score. Note that the area is calculated on the 2D image, since the sampled triangles in the 3D space could be very large due to depth variation, leading to unreasonable overestimation. 

\noindent 
Finally, the normal for a point $P_i$ is determined by a weighted combination of all the $K$ sampled candidates, where the weights represent the confidence given by our adaptive strategy:
\begin{equation}
\label{full_adaption}
\vec{n}_i=\frac{\sum_{k=0}^{K-1} s_{k} \cdot g_k \cdot \vec{n}_{k}}{\sum_{k=0}^{K-1} s_{k} \cdot g_k},
\end{equation}
where $K$ is the number of sampled triplets, $s_k$ is the projected area of three sampled point $(P_k^{A}, P_k^{B}, P_k^C)$ on the 2D image, and $n_k$ is its normal vector.

%% file: 4_implementation.tex
\section{Implementation}

\paragraph{Network architecture} Our network adopts a multi-scale structure, which consists of one encoder and two decoders. We use HRNet-48~\cite{wang2020deep} as our backbone.  Taking one image as input, one encoder produces coarse-to-fine estimated depths in four scales, and the other decoder is used to generate the guidance feature map that captures geometric context. The depth estimation decoder consists of four blocks in different scales, each of which is constituted by two ResNet~\cite{he2016deep} basic blocks with a feature reducing convolution layer. The appending convolution layers are used to produce one-channel output for depth estimation. The guidance feature encoder adopts a nearly identical structure with the depth encoder but without any feature reducing layers.

\vspace{-3mm}
\paragraph{Loss functions} Our training loss has two types of terms: depth loss term and surface normal loss term. 
For the depth term, we use the $L_1$ loss for our multi-scale estimation:
\begin{equation}
\label{depth_term}
    l_{d} = \sum_{s=0}^{3} \lambda^{s-3} \left\| D_{pred}^{s}-D_{gt}\right\|_{1},
\end{equation}
where $D_{pred}^{s}$ means the estimated depth map at $s^{th}$ scale, $D_{gt}$ is the ground truth depth map, and $\lambda$ is a factor for balancing different scales. Here we set $\lambda = 0.8$.

To enforce geometric constraint on the estimated depth map, using our proposed adaptive strategy, we compute the surface normals only based on the finest estimated depth map. To regularize the consistency of the computed surface normals with ground truth, we adopt a cosine embedding loss:

\begin{equation}
\label{normal_term}
    l_{n} = 1-cos(N_{pred}, N_{gt}),
\end{equation}
where $N_{pred}$ is the surface normal map calculated from the finest estimated depth map, and $N_{gt}$ is the ground truth surface normal. Therefore, the overall loss is defined as:
\begin{equation}
\label{full_loss}
    l = l_{d} + \alpha l_{n},
\end{equation}
where $\alpha$ is set to 5 in all experiments, which is a trade-off parameter to make the two types of terms roughly of the same scale.

\vspace{-3mm}
\paragraph{Training details}
Our model is implemented by Pytorch with Adam optimizer ($init\_lr= 0.0001$, $\beta_{1}=0.9$, $\beta_{2}=0.999$, $weight\_decay = 0.00001$). 
The learning rate is polynomially decayed with polynomial power 0.9.
The model is trained with only depth loss term in the first 20 epochs, and then with depth and surface normal loss terms in the last 20 epochs.
The whole training is completed with 8 batches on four GeForce RTX 2080 Ti GPUs. We adopt $5 \times 5$ local patch and 40 sampling triplets in all experiments. 

\begin{figure*}[t]
\setlength{\abovecaptionskip}{1pt}
\setlength{\belowcaptionskip}{1pt}
    \centering
    \includegraphics[width=\linewidth]{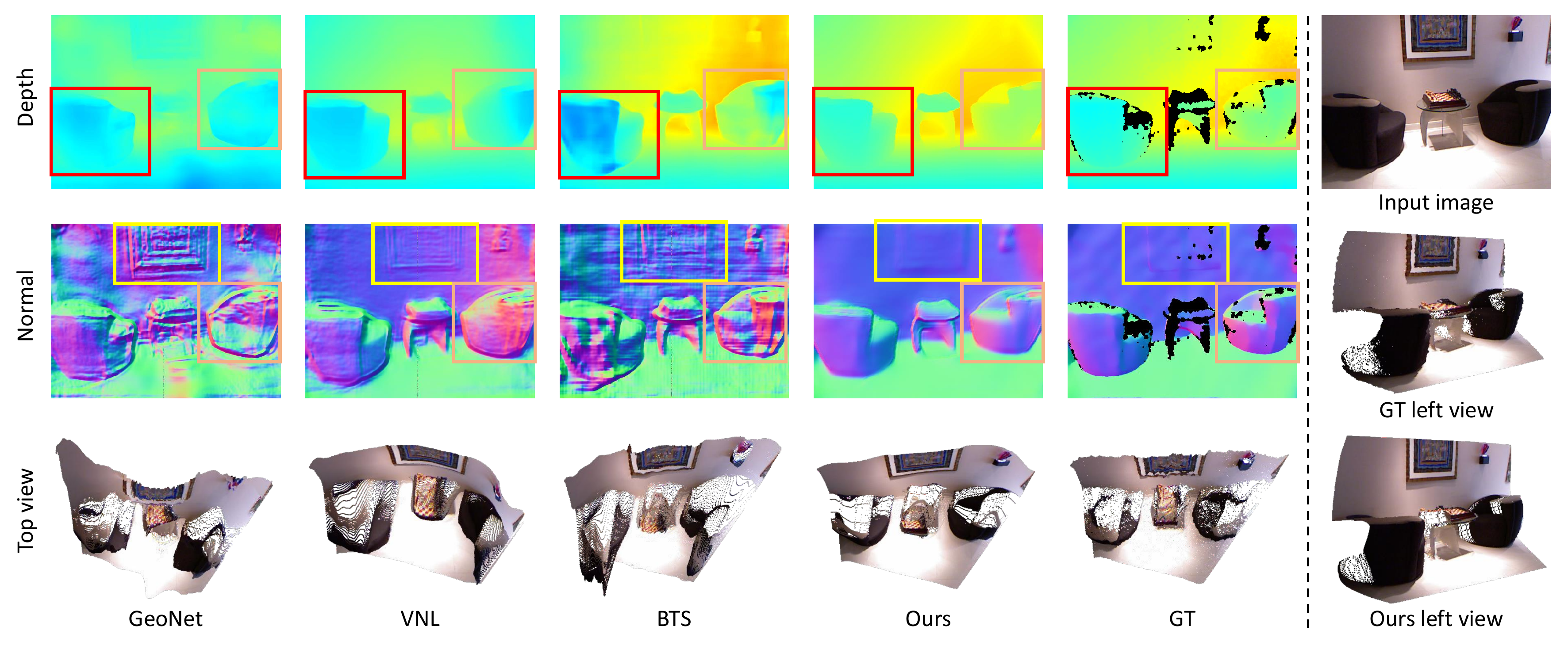}
    \caption{Qualitative comparisons with SOTAs on NYUD-V2. Compared with other methods, our estimated depth is more accurate and contain less noises. The recovered surface normal maps and point clouds demonstrate that our estimated depth faithfully preserve important geometric features. The black regions are the invalid regions lacking ground truth. }
    \label{fig:nyu_compare}
    \vspace{-3mm}
\end{figure*}

%% file: 5_0_experiments.tex
\section{Experiments}

\subsection{Dataset}
\paragraph{NYUD-V2} Our model is trained on NYUD-V2 dataset. NYUD-V2 is a widely used indoor dataset and contains 464 scenes, of which 249 scenes are for training and 215 for testing. We directly adopt the collected training data provided by Qi~\etal~\cite{qi2018geonet}, which has 30,816 frames sampled from the raw training scenes with precomputed ground truth surface normals. The precomputed surface normals are generated following the procedure of ~\cite{fouhey2013data}. Note that DORN~\cite{fu2018deep}, Eigen~\etal~\cite{eigen2015predicting}, Xu~\etal~\cite{xu2017multi}, Laina~\etal~\cite{laina2016deeper}, and Hu~\etal~\cite{hu2019revisiting} use 407k, 120k, 90k, 95k and 51k images for training, which are all significantly larger than ours. For testing, we utilize the official test set that is the same as the competitive methods, which contains 654 images.

\vspace{-5mm}
\paragraph{ScanNet} We also evaluate our method on a recently proposed indoor dataset, ScanNet~\cite{dai2017scannet}, which has more than 1600 scenes. 
Its official test split contains 100 scenes, and we uniformly select 2167 images from them for cross-dataset evaluation.

\begin{table}[t]
\setlength{\abovecaptionskip}{0pt}
\setlength{\belowcaptionskip}{0pt}
\begin{center}
\caption{Depth evaluation on NYUD-V2 dataset.
}
\label{tab:nyu_depth_eval}
\resizebox{\linewidth}{!}{%
\begin{tabular}{l | c c c c c c c}
\hline
Method & \textbf{rel} ($\downarrow$) & \textbf{log10} ($\downarrow$) & \textbf{rms} ($\downarrow$) & $ \boldsymbol{\delta_{1}}$ ($\uparrow$) & $ \boldsymbol{\delta_{2}}$ ($\uparrow$) &  $\boldsymbol{\delta_{3}}$ ($\uparrow$)\\
\hline
Saxena~\etal~\cite{saxena2008make3d} & 0.349 & - & 1.214 & 0.447 & 0.745 & 0.897\\ 
Karsch~\etal~\cite{karsch2012depth} & 0.349 & 0.131 & 1.21 & - & - & - \\
Liu~\etal~\cite{liu2014discrete} & 0.335 & 0.127 & 1.06 & - & - & - \\
Ladicky~\etal~\cite{ladicky2014pulling} & - & - & - & 0.542 & 0.829 & 0.941 \\
Li~\etal~\cite{li2015depth} & 0.232 & 0.094 & 0.821 & 0.621 & 0.886 & 0.968 \\
Roy~\etal~\cite{roy2016monocular} & 0.187 & 0.078 & 0.744 & - & - & - \\
Liu~\etal~\cite{liu2015learning} & 0.213 & 0.087 & 0.759 & 0.650 & 0.906 & 0.974 \\
Wang~\etal~\cite{wang2015towards} & 0.220 & 0.094 & 0.745 & 0.605 & 0.890 & 0.970\\
Eigen~\etal~\cite{eigen2015predicting} & 0.158 & - & 0.641 & 0.769 & 0.950 & 0.988 \\
Chakrabarti~\etal~\cite{chakrabarti2016depth} & 0.149 & - & 0.620 & 0.806 & 0.958 & 0.987 \\
Li~\etal~\cite{li2017two} & 0.143 & 0.063 & 0.635 & 0.788 & 0.958 & 0.991 \\
Laina~\etal~\cite{laina2016deeper} & 0.127 & 0.055 & 0.573 & 0.811 & 0.953 & 0.988 \\
Hu~\etal~\cite{hu2019revisiting} & 0.115 & 0.050 & 0.530 & 0.866 & 0.975 & 0.993 \\
DORN~\cite{fu2018deep} & 0.115 & 0.051 & 0.509 & 0.828 & 0.965 & 0.992 \\
GeoNet~\cite{qi2018geonet} & 0.128 &  0.057 & 0.569 & 0.834 & 0.960 & 0.990 \\
VNL~\cite{yin2019enforcing} & 0.108 & 0.048 & 0.416 & 0.875 & 0.976 & 0.994 \\
BTS~\cite{lee2019big} & 0.113 & 0.049 & 0.407 & 0.871 & 0.977 & 0.995 \\
Ours & \textbf{0.101} & \textbf{0.044} & \textbf{0.377} & \textbf{0.890} & \textbf{0.982} & \textbf{0.996}\\
\hline

\hline
\end{tabular}%
}
\end{center}
\vspace{-5mm}
\end{table}

\subsection{Evaluation metrics}
To evaluate our method, we compare our method with state-of-the-arts in three
aspects: accuracy of depth estimation, accuracy of recovered surface normal, and quality of recovered point cloud.

\vspace{-5mm}
\paragraph{Depth} Following the previous method~\cite{eigen2014depth}, we adopt the following metrics: mean absolute relative error (rel),
mean log10 error (log10), root mean squared error (rms), and the accuracy
under threshold ($\delta < {1.25}^{i} \text{ where } i\in\{1,2,3\}$).

\vspace{-5mm}
\paragraph{Surface normal} Like prior works~\cite{eigen2015predicting,qi2018geonet}, we evaluate surface normal with the following metrics: the mean of angle error (mean), the median of the angle error (median), and the accuracy below threshold $t \text{ where } t \in\left[11.25^{\circ}, 22.5^{\circ}, 30^{\circ}\right]$.

\vspace{-5mm}
\paragraph{Point cloud} 
For quantitatively evaluate the point clouds converted from estimated depth maps, we utilize the following metrics: mean Euclidean distance (dist), root mean squared Euclidean distance (rms), and the accuracy
below threshold $t \text{ where } t \in\left[0.1m, 0.3m, 0.5m\right]$.

\subsection{Evaluations}

\paragraph{Depth estimation accuracy} We compare our method with other state-of-the-art methods on NYUD-V2 dataset. As shown in Table~\ref{tab:nyu_depth_eval}, our method significantly outperforms the other SOTA methods across all evaluation metrics. Moreover, to further evaluate the generalization of our method, we compare our method with some strong SOTAs on unseen ScanNet dataset. As shown in Table~\ref{tab:scannet_depth_eval}, our method still shows better performance than others. 

Besides the quantitative comparison, we show some qualitative results for several SOTA methods that also use geometric constraints, including i) GeoNet~\cite{qi2018geonet} (least square normal); ii)VNL~\cite{yin2019enforcing} (virtual normal constraint); iii) BTS~\cite{lee2019big} (predict local plane equations not directly predict depth). As shown in Fig.~\ref{fig:nyu_compare}, the proposed method faithfully recovers the original geometry. For the regions with high curvatures, such as the sofas, our results obtain cleaner and smoother surfaces; our predicted depth map also yields high-quality shape boundaries, which leads to better accuracy compared to the ground truth depth map. Also, note even for the texture-less walls and floors, our estimated depth is still satisfactory.

\begin{table}[t]
\begin{center}
\caption{Depth evaluation on ScanNet dataset.
}
\label{tab:scannet_depth_eval}
\resizebox{\linewidth}{!}{%
\begin{tabular}{l | c c c c c c } %| c c c c c}
\hline
% \multirow{2}{*}{Method} & \textbf{rel} & \textbf{log10} & \textbf{rms} & $\delta_{1}$ & $\delta_{2}$ & $\delta_{3}$ & \textbf{Mean} & \textbf{Median} & $11.25^{\circ}$ & $22.5^{\circ}$ & $30^{\circ}$\\
%& \multicolumn{6}{c}{Depth evaluation} & \multicolumn{5}{|c}{Calculated surface normal evaluation} \\
Method & \textbf{rel} ($\downarrow$) & \textbf{log10} ($\downarrow$) & \textbf{rms} ($\downarrow$) & $ \boldsymbol{\delta_{1}}$ ($\uparrow$) & $ \boldsymbol{\delta_{2}}$ ($\uparrow$) &  $\boldsymbol{\delta_{3}}$ ($\uparrow$)\\
\hline
GeoNet~\cite{qi2018geonet} & 0.255 & 0.106 & 0.519 & 0.561 & 0.855 & \textbf{0.958}\\
VNL~\cite{yin2019enforcing}  & 0.238 & 0.105 & 0.505 & 0.565 & 0.856 & 0.957\\
% Hu~\etal~\cite{hu2019revisiting} (51k images)\\
BTS~\cite{lee2019big} & 0.246 & 0.104 & 0.506 & 0.583 & 0.858 & 0.951\\
Ours & \textbf{0.233} & \textbf{0.100} & \textbf{0.484} & \textbf{0.609} & \textbf{0.861} & 0.955 \\
\hline

\hline
\end{tabular}%
}
\end{center}
\vspace{-5mm}
\end{table}

\begin{table}[t]
\begin{center}
\caption{Point cloud evaluation on NYUD-V2 dataset.
}
\label{tab:point_cloud_eval_nyud}
\resizebox{\linewidth}{!}{%
\begin{tabular}{l | c c c c c } 
\hline
Method & \textbf{dist} ($\downarrow$) & \textbf{rms} ($\downarrow$) & $\boldsymbol{0.1m}$ ($\uparrow$) & $\boldsymbol{0.3m}$ ($\uparrow$) & $\boldsymbol{0.5m}$ ($\uparrow$) \\
\hline
VNL~\cite{yin2019enforcing} & 0.515 & 0.686 & 0.181 & 0.469 & 0.644 \\
GeoNet~\cite{qi2018geonet} & 0.392 & 0.608 & 0.220 & 0.558 & 0.747 \\
BTS~\cite{lee2019big}& 0.317 & 0.544 & 0.278 & 0.653 & 0.822 \\
Hu~\etal~\cite{hu2019revisiting} & 0.311 & 0.537 & 0.288 & 0.666 & 0.831\\
Ours & \textbf{0.266} & \textbf{0.497} & \textbf{0.332} & \textbf{0.727} & \textbf{0.869}\\
\hline

\hline
\end{tabular}%
}
\end{center}
\vspace{-5mm}
\end{table}

\vspace{-3mm}
\paragraph{Point cloud}
From the Table~\ref{tab:point_cloud_eval_nyud}, in terms of the quality of point cloud, our method outperforms other methods by a large margin. Surprisingly, although VNL~\cite{yin2019enforcing} has better performance than GeoNet~\cite{qi2018geonet} in terms of depth evaluation errors, its mean Eucleadian distance is worse than GeoNet, which demonstrates the necessity of evaluation specially designed for point clouds. As shown in Fig.~\ref{fig:nyu_compare} (the third row), our point cloud has fewer distortions and much more accurate than others. The point clouds generated from the depth maps of other methods suffer from severe distortions and struggle to preserve prominent geometric features, such as planes (e.g., walls) and surfaces with high curvatures (e.g., sofas). Besides, we also show a qualitative comparison between our point cloud and the ground truth from a different view in Fig.~\ref{fig:nyu_compare}. The highly consistent result further demonstrates our method's superior performance in terms of the quality of 3D geometry.

\vspace{-3mm}
\paragraph{Surface Normal}
As shown in Table~\ref{tab:nyu_normal_eval}, our recovered surface normals have considerably better quality than that of the other methods. For reference, we also report the results generated by the methods that directly output normal maps in the network. Surprisingly, the accuracy of our recovered surface normals is even higher than this kind of methods that can explicitly predict normals. Also, we present qualitative comparisons in Fig.~\ref{fig:nyu_compare}. It can be seen that our surface normal is smoother and more accurate than the others, which indicates that our strategy is effective for correlating normal constraints with depth estimation, resulting in not only accurate depth estimation, but also reliable surface normals and 3D geometry.

\begin{table}[t]
\begin{center}
\caption{Surface normal evaluation on NYUD-V2 dataset.
}
\label{tab:nyu_normal_eval}
\resizebox{\linewidth}{!}{%
\begin{tabular}{l | c c c c c}
\hline
Method & \textbf{Mean} ($\downarrow$) & \textbf{Median} ($\downarrow$) & $\boldsymbol{11.25^{\circ}}$ ($\uparrow$) & $\boldsymbol{22.5^{\circ}}$ ($\uparrow$) & $\boldsymbol{30^{\circ}}$ ($\uparrow$)\\
\hline
\hline
\multicolumn{6}{c}{Predicted Surface Normal from the Network} \\
\hline
\hline
3DP~\cite{fouhey2013data} & 33.0 & 28.3 & 18.8 & 40.7 & 52.4 \\
Ladicky~\etal~\cite{ladicky2014pulling} & 35.5 & 25.5 & 24.0 & 45.6 & 55.9 \\
Fouhey~\etal~\cite{fouhey2014unfolding} & 35.2 & 17.9 & 40.5 & 54.1 & 58.9 \\
Wang~\etal~\cite{wang2015designing} & 28.8 & 17.9 & 35.2 & 57.1 & 65.5 \\
Eigen~\etal~\cite{eigen2015predicting} & 23.7 & 15.5 & 39.2 & 62.0 & 71.1 \\
\hline
\hline
\multicolumn{6}{c}{Calculated Surface Normal from the Point cloud} \\
\hline
\hline
BTS~\cite{lee2019big} & 44.0 & 35.4 & 14.4 & 32.5 & 43.2 \\
GeoNet~\cite{qi2018geonet} & 36.8 & 32.1 & 15.0 & 34.5 & 46.7 \\
DORN~\cite{fu2018deep} & 36.6 & 31.1 & 15.7 & 36.5 & 49.4  \\
Hu~\etal~\cite{hu2019revisiting} & 32.1 & 23.5 & 24.7 & 48.4 & 59.9 \\
VNL~\cite{yin2019enforcing} & 24.6 & 17.9 & 34.1 & 60.7 & 71.7  \\
Ours & \textbf{20.0} & \textbf{13.4} & \textbf{43.5} & \textbf{69.1} & \textbf{78.6} \\
\hline

\hline
\end{tabular}%
}
\end{center}
\vspace{-5mm}
\end{table}

% Firstly, we review the pinhole camera model. For any pixel $(u,v)$ of the depth map, its corresponding 3D point $(X,Y,Z)$ is:
% \begin{equation}
% \label{camera_model}
% \begin{aligned}
% X &=\left(u-c_{x}\right) * z / f_{x} \\
% Y &=\left(v-c_{y}\right) * z / f_{y} \\
% Z &= z
% \end{aligned}
% \end{equation}
% where $(c_x, c_y)$ is the principle point and $fx,fy$ are the focal lengths along X-axis and Y-axis respectively. \LXX{remove this part. shorter}

% \LXX{Assuming the estimated depth $z$ and ground truth depth $\hat{z}$ has a error $\Delta z$, based on pinhole camera model, the Euclidean distance of the predicted 3D point (X,Y,Z) and ground truth point $(\hat{X}, \hat{Y}, \hat{Z})$ can be obtained:
% \begin{equation}
% \label{Euclidean_dist}
% \begin{aligned}
% dist &=\sqrt{(\Delta X)^{2}+(\Delta Y)^{2}+(\Delta Z)^{2}} \\
% &=\sqrt{\frac{(u-C_{x})^{2}}{f_{x}^{2}}(\Delta z)^{2}+\frac{(v-C_{y})^{2}}{f_{y}^{2}}(\Delta z)^{2}+(\Delta z)^{2}} \\
% &=\sqrt{\frac{(u-C_{x})^{2}}{f_{x}^{2}}+\frac{\left(v-C_{y}\right)^{2}}{f_{y}^{2}}+1} \cdot \Delta z .
% \end{aligned}
% \end{equation}
% From the Equation~\ref{Euclidean_dist}, we can know the pixels far from the camera center will have larger euclidean distances while the near ones will have relative smaller distances, even if they have similar depth errors.(move to supp.)}

%% file: 5_1_discussions.tex
\subsection{Discussions}
In this section, we further conduct a series of evaluations with an HRNet-18~\cite{wang2020deep} backbone to give more insights into the proposed method.

\vspace{-3mm}
\paragraph{Effectiveness of ASN}
To validate the effectiveness of our proposed adaptive surface normal constraint, we train models with different constraints: a) only $L_1$ depth constraint; b) depth and Sobel-like operator surface normal constraints (SOSN); c) depth and least square surface normal constraints (LSSN); d) depth and virtual normal constraints (VN); e) depth and our adaptive surface normal constraints (ASN).

% e) depth and our geometric-context aware surface normal constraints; f) depth and our pixel adaptive surface normal constraints.

As shown in Table~\ref{tab:diff_constraints}, the model with our adaptive surface normal constraint outperforms (ASN) all the others. Although the models with Sobel-like operator (SOSN) and least square normal constraint (LSSN) shows better recovered surface normal, their depth estimation accuracy drops off compared with the model without geometric constraint. The model with virtual normal (VN)~\cite{yin2019enforcing} constraint shows the worst quality of recovered surface normal among the four types of geometric constraints, given that virtual normal is derived from global sampling on the estimated depth map, which inevitably loses local geometric information.  

\begin{figure}[t]
\setlength{\abovecaptionskip}{1pt}
\setlength{\belowcaptionskip}{1pt}
    \centering
    \includegraphics[width=\linewidth]{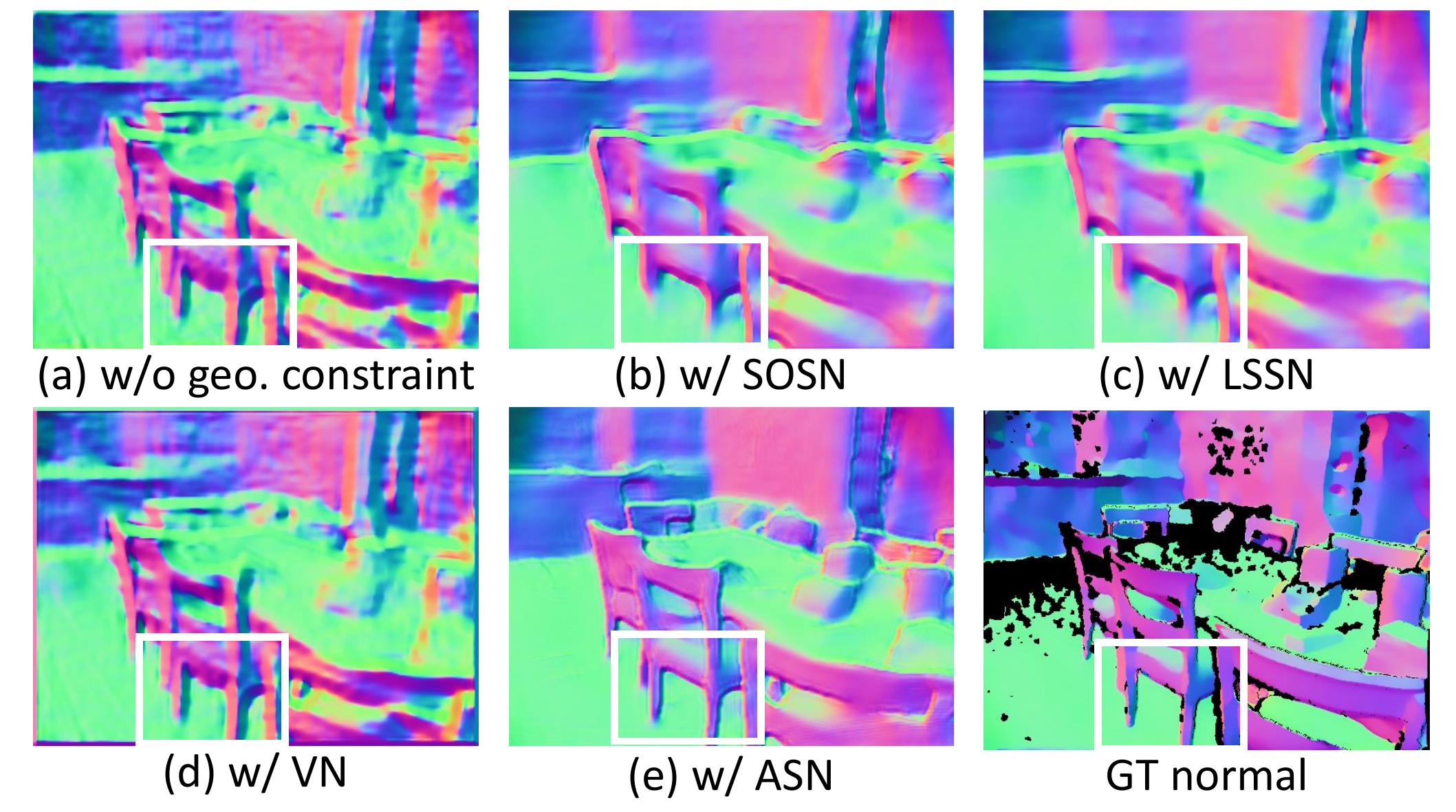}
    \caption{Comparisons of models with different geometric constraints. Model with our ASN constraint achieves better surface normal estimation, even accurately capture detailed geometries, like the legs of chairs (see white boxes).}
    \label{fig:different_constraints}
\end{figure}

Furthermore, we give a set of qualitative comparisons in Fig.~\ref{fig:different_constraints}. The results clearly show our ASN constraint achieves better surface normal estimation results and captures detailed geometric features, even for the thin structures like the legs of chairs.

\begin{table}[t]
\begin{center}
\caption{Comparisons of models with different geometric constraints on NYUD-V2 dataset.
}
\label{tab:diff_constraints}
\resizebox{\linewidth}{!}{%
\begin{tabular}{l | c c c | c c c }
\hline
\multirow{2}{*}{Constraints} & \textbf{rel} ($\downarrow$) & \textbf{log10} ($\downarrow$) & $\boldsymbol{\delta_{1}}$ ($\uparrow$) & \textbf{Mean} ($\downarrow$) & \textbf{Median} ($\downarrow$) & $\boldsymbol{11.25^{\circ}}$ ($\uparrow$) \\
& \multicolumn{3}{c}{Depth} & \multicolumn{3}{|c}{Recovered normal} \\
\hline
L1 & 0.113 & 0.047  & 0.875   & 31.3 & 23.2 & 24.9 \\
L1 + SOSN & 0.118 & 0.049 & 0.867  & 22.8 & 16.1 & 36.2 \\
L1 + LSSN & 0.119 & 0.050  & 0.862  & 23.5 & 16.3 & 35.7 \\
L1 + VN & 0.111 & 0.047  & 0.876  & 31.7 & 21.4 & 28.4 \\
% L1 + GASN & 0.112 & 0.047 & 0.876 & 22.4 & 15.9 & 36.8 \\
L1 + ASN & \textbf{0.111} & \textbf{0.047} & \textbf{0.876} & \textbf{22.2} & \textbf{15.8} & \textbf{36.9} \\
\hline

\hline
\end{tabular}%
}
\end{center}
\vspace{-4mm}
\end{table}

\vspace{-3mm}
\paragraph{Ablation study of adaptive modules}
To evaluate the effect of the proposed two adaptive modules, i.e., geometric context adaption and area adaption, we conduct an ablation study. We train models with different adaptive modules: only Geometric Context (GC) adaption, only Area adaption, and both. From Table~\ref{tab:abalation}, we can see the model with both adaptive modules achieves the best performance, which verifies the necessity of each adaptive module. 

\begin{table}[t]
\begin{center}
\caption{Ablation study of the proposed adaptive modules on NYUD-V2 dataset. We evaluate the accuracy of the recovered surface normals.}
\label{tab:abalation}
\resizebox{\linewidth}{!}{%
\begin{tabular}{l | c c c c c }
\hline
Module & \textbf{Mean} ($\downarrow$) & \textbf{Median} ($\downarrow$) & $\boldsymbol{11.25^{\circ}}$ ($\uparrow$) & $\boldsymbol{22.5^{\circ}}$ ($\uparrow$) & $\boldsymbol{30^{\circ}}$ ($\uparrow$)\\
\hline
only Area &22.6 & 16.0 & 36.4 & 63.6 & 74.4 \\
only GC & 22.3 & \textbf{15.8} & \textbf{36.9} & 64.1 & 74.8\\
Area+GC & \textbf{22.2} & \textbf{15.8} & \textbf{36.9} & \textbf{64.2} & \textbf{74.9} \\
\hline

\hline
\end{tabular}%
}
\end{center}
\vspace{-4mm}
\end{table}

\begin{figure}[t]
\setlength{\abovecaptionskip}{1pt}
\setlength{\belowcaptionskip}{1pt}
    \centering
    \includegraphics[width=\linewidth]{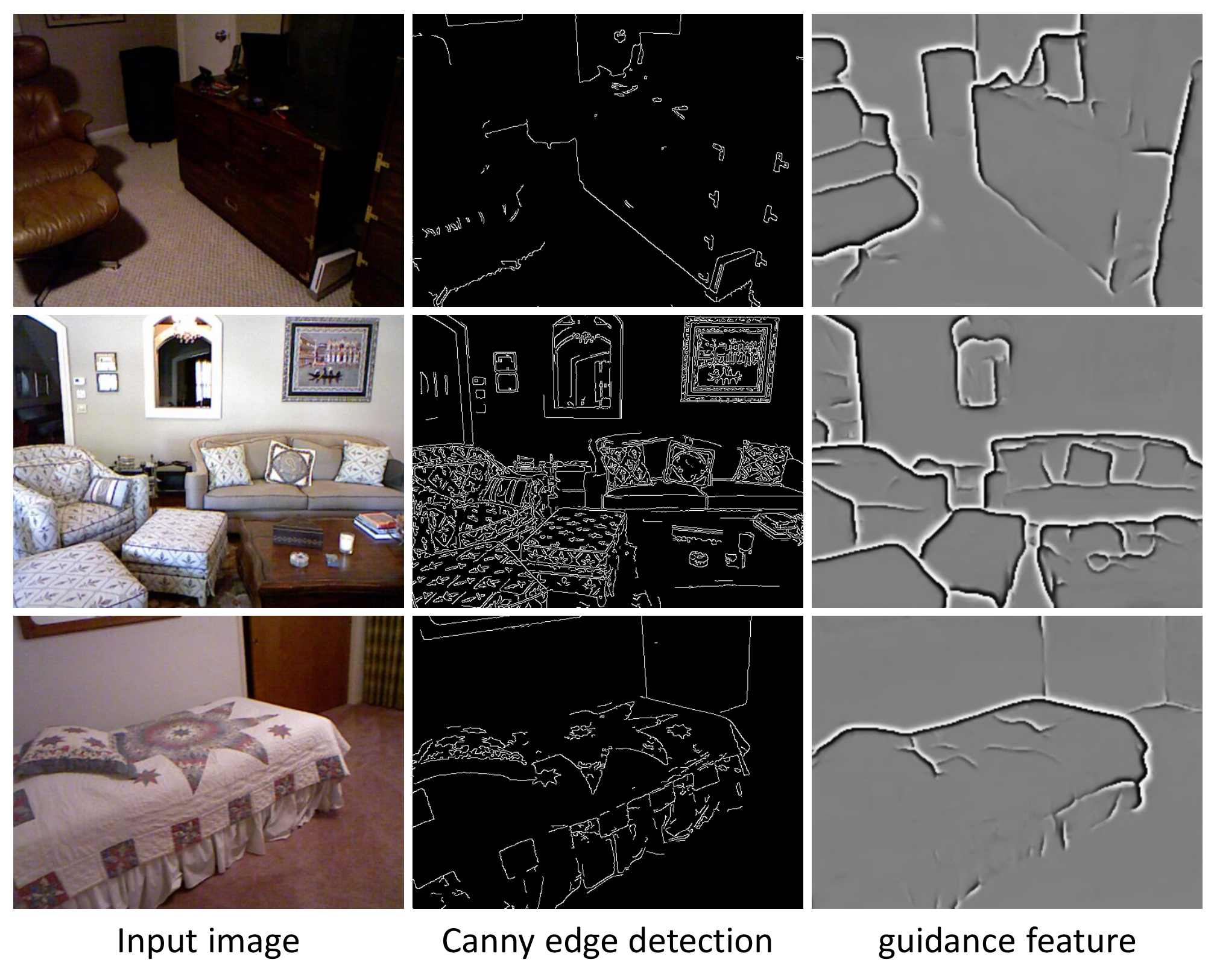}
    \caption{Our guidance feature maps versus edge maps detected by Canny operator. Although shape boundaries have high statistic correlations with image edges, they are not always coincident. Our feature map captures the true geometric boundaries, while the Canny operator detects edges with significant intensity variances.}
    \label{fig:guidance_feature_vis}
\end{figure}

\paragraph{Visualization of guidance features} 
The geometric adaptive module is the key to our adaptive surface normal constraint method. To better understand what the network learns, we visualize the learned features of the guidance map. We plot one from the multiple channels of the guidance feature map, which is shown in Fig.~\ref{fig:guidance_feature_vis}. It can be seen that the learned guidance map captures the shape context and geometric variations, giving informative boundaries. 

For comparison, we use the Canny operator to detect the edges of the input image by image intensity variance. As we can see, our guidance feature map is not simply coincident with the Canny edges. For example, in Fig.~\ref{fig:guidance_feature_vis}, the Canny operator detects fragmented edges based on the texture of wall painting and sofas, while our guidance feature map indicates the true shape boundaries where the 3D geometry changes.

\begin{figure}[t]
\setlength{\abovecaptionskip}{1pt}
\setlength{\belowcaptionskip}{1pt}
    \centering
    \includegraphics[width=\linewidth]{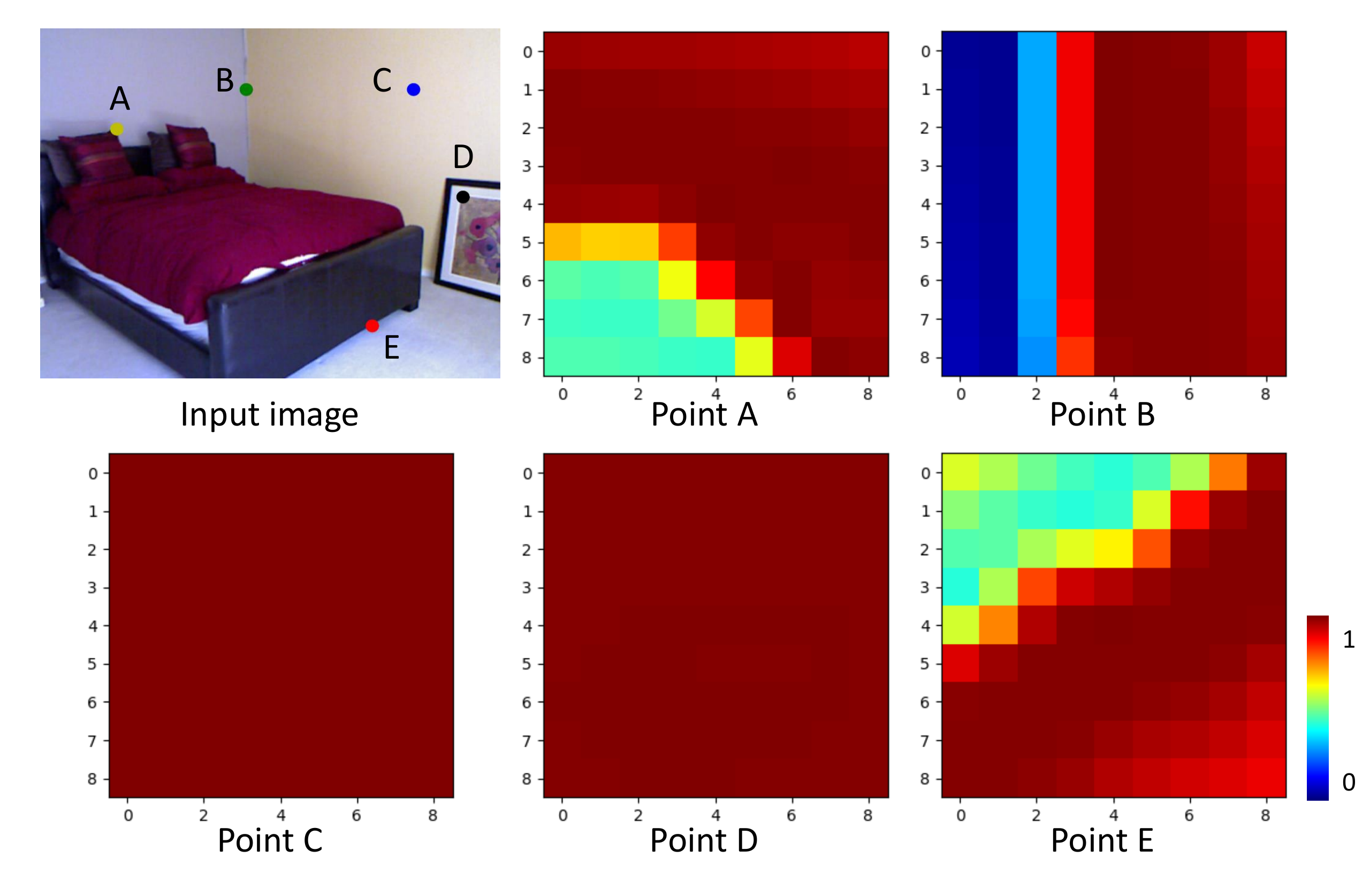}
    \caption{The visualization of similarity kernels. 
    The similarity kernels of Point A, B, and E demonstrate that our method could successfully distinguish different geometries. The similarity kernels of Point C and D further show that our method captures the 3D geometry variances of the shapes in the 3D world, instead of the image color distinctions. 
    }
    \label{fig:kernel_vis}
\end{figure}

\paragraph{Visualization of similarity kernels}
To validate whether our model can capture the true geometric boundaries of shapes, we select five points on the image and visualize their color coded similarity kernels in Fig.~\ref{fig:kernel_vis}. The similarity kernels of Point A, B, and E indicate that our method could successfully distinguish different geometries, such as shape boundaries and corners. Furthermore, the similarity kernels of Point C and D show that our approach captures the 3D geometry variances of the shapes in the real world, instead of the color distinctions of the image. For example,  Point D has large color variances in the image, but its similarity kernel has constant values indicating the unchanged geometry.

\begin{figure}[t]
\setlength{\abovecaptionskip}{1pt}
\setlength{\belowcaptionskip}{1pt}
    \centering
    \includegraphics[width=\linewidth]{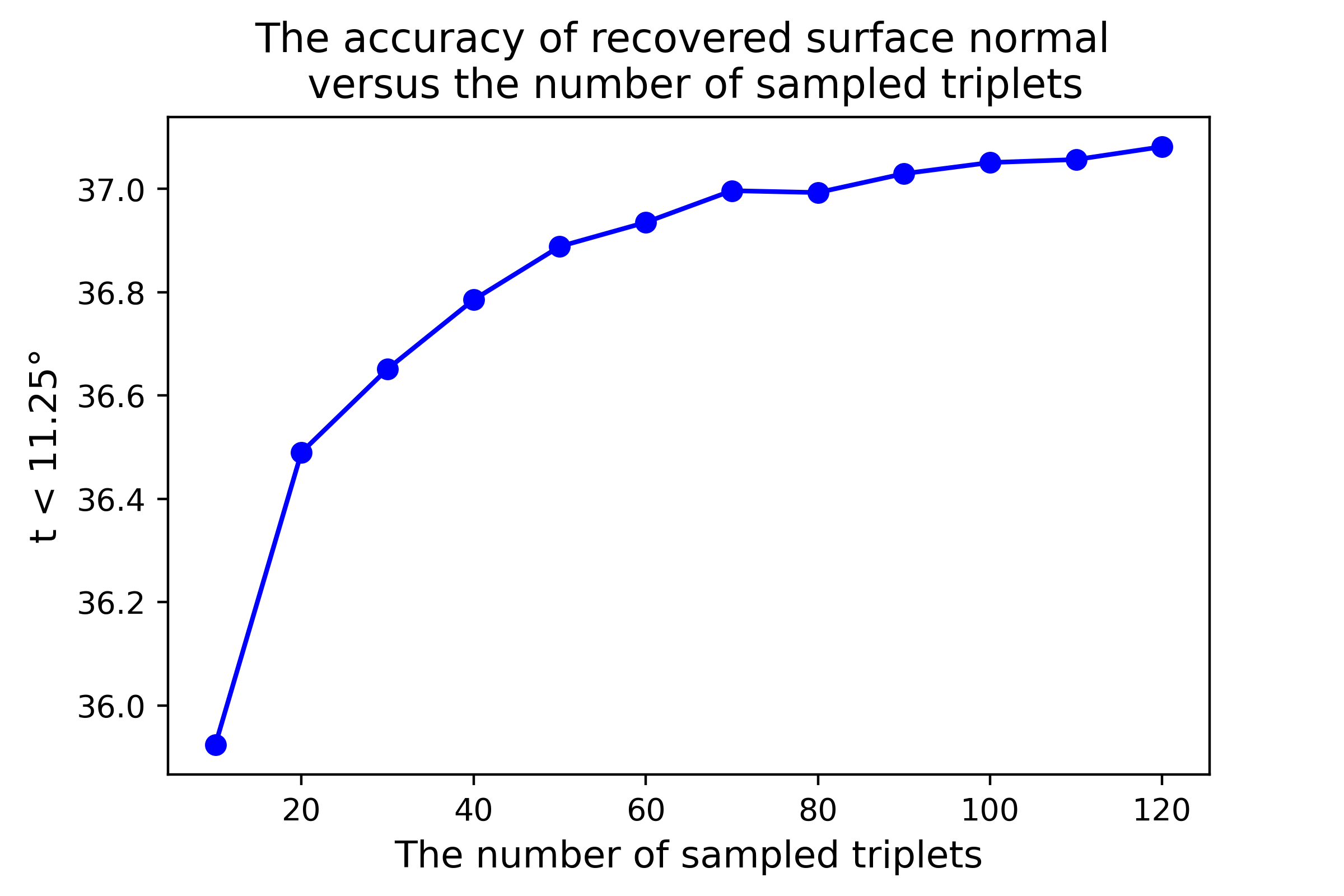}
    \caption{The accuracy of recovered surface normal versus the number of sampled triplets. The more triplets are sampled, the more accurate the recovered surface normal is.}
    \label{fig:num_of_triplets}
\end{figure}

\paragraph{Number of sampled triplets.}
To quantitatively analyze the influence of the number of sampled triplets, we recover surface normals from our estimated depth maps using our adaptive surface normal computation method with $5 \times 5$ local patch.
Based on Fig.~\ref{fig:num_of_triplets}, it is not surprised that more sampled triplets will contribute to more accurate surface normals. The accuracy increases dramatically from $10 \sim 20$ sampled triplets and gradually saturates with more triplets sampled. To balance efficiency and accuracy, the number of sampled triplets is recommended to be $40 \sim 60$.

\paragraph{Size of local patch.}
We evaluate the effect of the size of local patch to our method by training the network with different local patch sizes. As illustrated in Table~\ref{tab:patch_size}, a larger local patch could improve the performance, especially for the surface normal, but the improvements are not significant. The reason behind is, our ASN constraint is an adaptive strategy that can automatically determine the reliability of the sampled points given different local patches; therefore, our method is robust to the choice of local patch size.

\begin{table}[t]
\begin{center}
\caption{The influence of local patch size.
}
\label{tab:patch_size}
\resizebox{0.95\linewidth}{!}{%
\begin{tabular}{c | c c c  | c c c}
\hline
\multirow{2}{*}{Size} & \textbf{rel} ($\downarrow$) & \textbf{log10} ($\downarrow$) & $\boldsymbol{\delta_{1}}$ ($\uparrow$) & \textbf{Mean} ($\downarrow$) & \textbf{Median} ($\downarrow$) & $\boldsymbol{11.25^{\circ}}$ ($\uparrow$) \\
& \multicolumn{3}{c}{Depth} & \multicolumn{3}{|c}{ Recovered Normal} \\
\hline
3 & 0.112 & 0.047 & \textbf{0.877} & 22.5 & 15.8 & 36.9\\
5 & \textbf{0.111} & 0.047 & 0.876 & 22.4 & 15.8 & \textbf{37.1}\\
7 & 0.112 & 0.047 & \textbf{0.877} & \textbf{22.2} & \textbf{15.7} & \textbf{37.1}\\
9 & \textbf{0.111} & 0.047 & 0.875 & 22.4 & 15.8 & 37.0 \\
\hline
\end{tabular}%
}
\end{center}
\vspace{-4mm}
\end{table}

\begin{figure}[t]
\setlength{\abovecaptionskip}{1pt}
\setlength{\belowcaptionskip}{1pt}
    \centering
    \includegraphics[width=\linewidth]{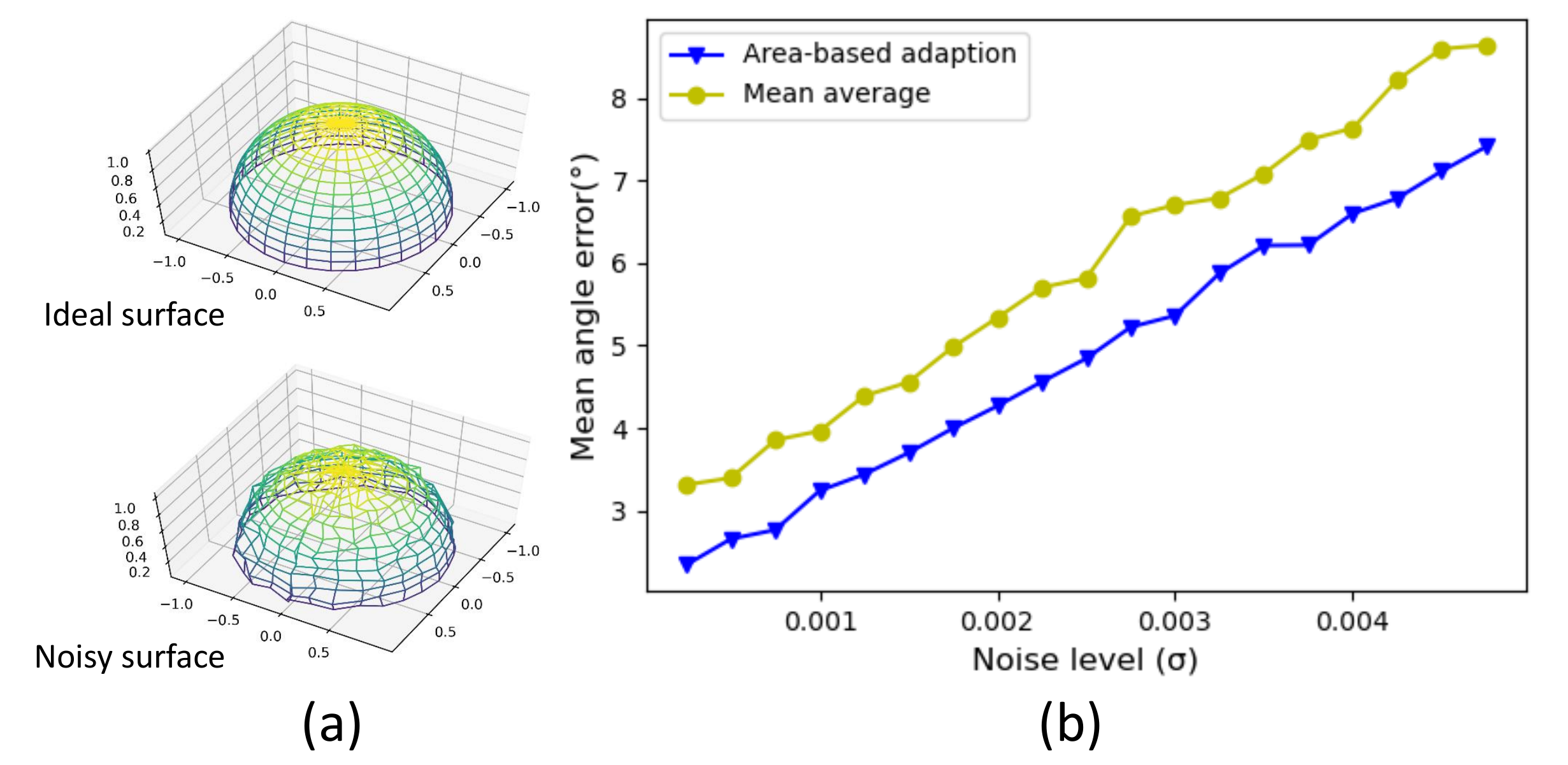}
    \caption{Effectiveness of the area-based adaption. (a) The ideal and noisy surface. (b) We employ the mean angle error to evaluate surface normals estimated by the simple average strategy and our area-based adaption. Compared with the simple average strategy, our area-based adaption is more robust to noises.}
    \label{fig:noise_exp}
\end{figure}

\paragraph{Area-based adaption} We use the area of a sampled triangle as the combinational weight for adaption. To evaluate the effectiveness of the area-based adaption, we conduct an experiment with a comparison to the simple average strategy. We create a unit semi-sphere surface as noise-free data and then add Gaussian noises to simulate real noisy data (see Fig.~\ref{fig:noise_exp} (a)). We compare the mean of angle errors of the normals estimated by these two methods with the increase of noises, and the results are given in Fig.~\ref{fig:noise_exp} (b). We can see that our area-based adaption gives lower estimation error with the increase of noise level, demonstrating the robustness of the use of area for adaption.

\paragraph{Time complexity}
Here, we discuss the time complexity of different normal computation methods, including our sampling based method, Sobel-like operator \cite{hu2019revisiting, kusupati2020normal} and least square based method \cite{qi2018geonet,long2020occlusion} .
Ours and the Sobel-like operator only involve matrix addition and vector dot/cross production operations; thus it is easy to show the time complexity is $O\left(n \right)$, while our time complexity will increase linearly with more samples. 
% Ours should be $O\left(m \cdot n \right)$, where $m$ is the number of sampled triplets, if $m=1$, our complexity should be $O\left(n \right)$. 
However, the least square module \cite{qi2018geonet,long2020occlusion} directly calculates the closed form solution of least square equations, which involves matrix multiplication, inversion and determinant, leading to the time complexity of $O\left(n^{3} \right)$.
Therefore, our method effectively balances the accuracy and computational efficiency.
% For example, to calculate the surface normal of a batch of images with size $H,W$, the computational resources will be huge to compute matrix multiplication and inversion for matrices with size $(b,h,w,k^{2},3)$ and $(b,h,w,k^{2},k^{2})$, where $b$ is the batch size and $k$ is the local patch size. With lager image size or larger local patch, the consuming memory will increase exponentially. Due to our method only involves matrix addition and inner/outer product, its consuming memory just increases linearly with larger local patch. For one image with size $640\times480$, the surface normal calculation will consumes 11GB GPU memories if adopting $9 \times 9$ local patch. 

% \begin{table*}[t]
% \begin{center}
% \caption{The effectiveness of pixel adaptive surface normal. on NYUD-V2 dataset.
% }
% \label{tab:diff_constraints}
% \resizebox{\linewidth}{!}{%
% \begin{tabular}{l | c c c c c c | c c c c c}
% \hline
% \multirow{2}{*}{Method} & \textbf{rel} & \textbf{log10} & \textbf{rms} & $\delta_{1}$ & $\delta_{2}$ & $\delta_{3}$ & \textbf{Mean} & \textbf{Median} & $11.25^{\circ}$ & $22.5^{\circ}$ & $30^{\circ}$\\
% & \multicolumn{6}{c}{Depth evaluation} & \multicolumn{5}{|c}{Calculated surface normal evaluation} \\
% \hline
% L1 & 0.113 & 0.047 & 0.393 & 0.875 & 0.978 & 0.996  & 31.3 & 23.2 & 24.9 & 48.8 & 60.4\\
% L1 + sobel normal & 0.118 & 0.049 & 0.405 & 0.867 & 0.975 & 0.995 & 22.8 & 16.1 & 36.2 & 63.6 & 74.3\\
% L1 + least square normal & 0.119 & 0.050 & 0.408 & 0.862 & 0.975 & 0.995 & 23.5 & 16.3 & 35.7 & 63.1 & 73.8\\
% L1 + virtual normal & 0.111 & 0.047 & 0.393 & 0.876 & 0.979 & 0.996 & 31.7 & 21.4 & 28.4 & 51.9 & 62.2\\
% L1 + sampling normal & 0.112 & 0.047 & 0.393 & 0.876 & 0.978 & 0.995 & 22.4 & 15.9 & 36.8 & 63.9 & 74.6\\
% L1 + pixel-adaptive sampling normal & 0.111 & 0.047 & 0.391 & 0.876 & 0.979 & 0.996 & 22.4 & 15.8 & 37.1 & 64.0 & 74.6\\
% \hline

% \hline
% \end{tabular}%
% }
% \end{center}
% \end{table*}

%% file: 6_conclusion.tex
\section{Conclusion}
In this paper, we present a simple but effective Adaptive Surface Normal (ASN) constraint for monocular depth estimation. 
Compared with other surface normal constraints, our constraint could adaptively determine the reliable local geometry for normal calculation, by jointly leveraging the latent image features and explicit geometry properties.
With the geometric constraint, our model not only estimates accurate depth maps but also faithfully preserves important 3D geometric features so that high-quality 3D point clouds and surface normal maps can be recovered from the estimated depth maps. 

\begin{acks}
We thank the anonymous reviewers for their valuable feedback. Wenping Wang acknowledges support from AIR@InnoHK -- Center for Transformative Garment Production (TransGP). Christian Theobalt acknowledges support from ERC Consolidator Grant 4DReply (770784). Lingjie Liu acknowledges support from Lise Meitner Postdoctoral Fellowship. 
\end{acks}

%% file: egpaper_for_review.bbl
\begin{thebibliography}{10}\itemsep=-1pt

\bibitem{bao2014fast}
Linchao Bao, Qingxiong Yang, and Hailin Jin.
\newblock Fast edge-preserving patchmatch for large displacement optical flow.
\newblock In {\em Proceedings of the IEEE Conference on Computer Vision and
  Pattern Recognition}, pages 3534--3541, 2014.

\bibitem{black1998robust}
Michael~J Black, Guillermo Sapiro, David~H Marimont, and David Heeger.
\newblock Robust anisotropic diffusion.
\newblock {\em IEEE Transactions on image processing}, 7(3):421--432, 1998.

\bibitem{chakrabarti2016depth}
Ayan Chakrabarti, Jingyu Shao, and Gregory Shakhnarovich.
\newblock Depth from a single image by harmonizing overcomplete local network
  predictions.
\newblock {\em arXiv preprint arXiv:1605.07081}, 2016.

\bibitem{dai2017scannet}
Angela Dai, Angel~X Chang, Manolis Savva, Maciej Halber, Thomas Funkhouser, and
  Matthias Nie{\ss}ner.
\newblock Scannet: Richly-annotated 3d reconstructions of indoor scenes.
\newblock In {\em Proceedings of the IEEE Conference on Computer Vision and
  Pattern Recognition}, pages 5828--5839, 2017.

\bibitem{eigen2015predicting}
David Eigen and Rob Fergus.
\newblock Predicting depth, surface normals and semantic labels with a common
  multi-scale convolutional architecture.
\newblock In {\em Proceedings of the IEEE international conference on computer
  vision}, pages 2650--2658, 2015.

\bibitem{eigen2014depth}
David Eigen, Christian Puhrsch, and Rob Fergus.
\newblock Depth map prediction from a single image using a multi-scale deep
  network.
\newblock {\em arXiv preprint arXiv:1406.2283}, 2014.

\bibitem{fouhey2013data}
David~F Fouhey, Abhinav Gupta, and Martial Hebert.
\newblock Data-driven 3d primitives for single image understanding.
\newblock In {\em Proceedings of the IEEE International Conference on Computer
  Vision}, pages 3392--3399, 2013.

\bibitem{fouhey2014unfolding}
David~Ford Fouhey, Abhinav Gupta, and Martial Hebert.
\newblock Unfolding an indoor origami world.
\newblock In {\em European Conference on Computer Vision}, pages 687--702.
  Springer, 2014.

\bibitem{fu2018deep}
Huan Fu, Mingming Gong, Chaohui Wang, Kayhan Batmanghelich, and Dacheng Tao.
\newblock Deep ordinal regression network for monocular depth estimation.
\newblock In {\em Proceedings of the IEEE Conference on Computer Vision and
  Pattern Recognition}, pages 2002--2011, 2018.

\bibitem{godard2017unsupervised}
Cl{\'e}ment Godard, Oisin Mac~Aodha, and Gabriel~J Brostow.
\newblock Unsupervised monocular depth estimation with left-right consistency.
\newblock In {\em Proceedings of the IEEE Conference on Computer Vision and
  Pattern Recognition}, pages 270--279, 2017.

\bibitem{godard2019digging}
Cl{\'e}ment Godard, Oisin Mac~Aodha, Michael Firman, and Gabriel~J Brostow.
\newblock Digging into self-supervised monocular depth estimation.
\newblock In {\em Proceedings of the IEEE/CVF International Conference on
  Computer Vision}, pages 3828--3838, 2019.

\bibitem{he2016deep}
Kaiming He, Xiangyu Zhang, Shaoqing Ren, and Jian Sun.
\newblock Deep residual learning for image recognition.
\newblock In {\em Proceedings of the IEEE conference on computer vision and
  pattern recognition}, pages 770--778, 2016.

\bibitem{hu2019revisiting}
Junjie Hu, Mete Ozay, Yan Zhang, and Takayuki Okatani.
\newblock Revisiting single image depth estimation: Toward higher resolution
  maps with accurate object boundaries.
\newblock In {\em 2019 IEEE Winter Conference on Applications of Computer
  Vision (WACV)}, pages 1043--1051. IEEE, 2019.

\bibitem{jiao2014local}
Jianbo Jiao, Ronggang Wang, Wenmin Wang, Shengfu Dong, Zhenyu Wang, and Wen
  Gao.
\newblock Local stereo matching with improved matching cost and disparity
  refinement.
\newblock {\em IEEE MultiMedia}, 21(4):16--27, 2014.

\bibitem{karsch2012depth}
Kevin Karsch, Ce Liu, and Sing~Bing Kang.
\newblock Depth extraction from video using non-parametric sampling.
\newblock In {\em European Conference on Computer Vision}, pages 775--788.
  Springer, 2012.

\bibitem{kusupati2020normal}
Uday Kusupati, Shuo Cheng, Rui Chen, and Hao Su.
\newblock Normal assisted stereo depth estimation.
\newblock In {\em Proceedings of the IEEE/CVF Conference on Computer Vision and
  Pattern Recognition}, pages 2189--2199, 2020.

\bibitem{ladicky2014pulling}
Lubor Ladicky, Jianbo Shi, and Marc Pollefeys.
\newblock Pulling things out of perspective.
\newblock In {\em Proceedings of the IEEE conference on computer vision and
  pattern recognition}, pages 89--96, 2014.

\bibitem{laina2016deeper}
Iro Laina, Christian Rupprecht, Vasileios Belagiannis, Federico Tombari, and
  Nassir Navab.
\newblock Deeper depth prediction with fully convolutional residual networks.
\newblock In {\em 2016 Fourth international conference on 3D vision (3DV)},
  pages 239--248. IEEE, 2016.

\bibitem{lee2019big}
Jin~Han Lee, Myung-Kyu Han, Dong~Wook Ko, and Il~Hong Suh.
\newblock From big to small: Multi-scale local planar guidance for monocular
  depth estimation.
\newblock {\em arXiv preprint arXiv:1907.10326}, 2019.

\bibitem{li2015depth}
Bo Li, Chunhua Shen, Yuchao Dai, Anton Van Den~Hengel, and Mingyi He.
\newblock Depth and surface normal estimation from monocular images using
  regression on deep features and hierarchical crfs.
\newblock In {\em Proceedings of the IEEE conference on computer vision and
  pattern recognition}, pages 1119--1127, 2015.

\bibitem{li2017two}
Jun Li, Reinhard Klein, and Angela Yao.
\newblock A two-streamed network for estimating fine-scaled depth maps from
  single rgb images.
\newblock In {\em Proceedings of the IEEE International Conference on Computer
  Vision}, pages 3372--3380, 2017.

\bibitem{liu2015learning}
Fayao Liu, Chunhua Shen, Guosheng Lin, and Ian Reid.
\newblock Learning depth from single monocular images using deep convolutional
  neural fields.
\newblock {\em IEEE transactions on pattern analysis and machine intelligence},
  38(10):2024--2039, 2015.

\bibitem{liu2021learning}
Lina Liu, Yiyi Liao, Yue Wang, Andreas Geiger, and Yong Liu.
\newblock Learning steering kernels for guided depth completion.
\newblock {\em IEEE Transactions on Image Processing}, 30:2850--2861, 2021.

\bibitem{liu2020fcfr}
Lina Liu, Xibin Song, Xiaoyang Lyu, Junwei Diao, Mengmeng Wang, Yong Liu, and
  Liangjun Zhang.
\newblock Fcfr-net: Feature fusion based coarse-to-fine residual learning for
  monocular depth completion.
\newblock {\em arXiv preprint arXiv:2012.08270}, 2020.

\bibitem{liu2014discrete}
Miaomiao Liu, Mathieu Salzmann, and Xuming He.
\newblock Discrete-continuous depth estimation from a single image.
\newblock In {\em Proceedings of the IEEE Conference on Computer Vision and
  Pattern Recognition}, pages 716--723, 2014.

\bibitem{long2020occlusion}
Xiaoxiao Long, Lingjie Liu, Christian Theobalt, and Wenping Wang.
\newblock Occlusion-aware depth estimation with adaptive normal constraints.
\newblock In {\em European Conference on Computer Vision}, pages 640--657.
  Springer, 2020.

\bibitem{perona1990scale}
Pietro Perona and Jitendra Malik.
\newblock Scale-space and edge detection using anisotropic diffusion.
\newblock {\em IEEE Transactions on pattern analysis and machine intelligence},
  12(7):629--639, 1990.

\bibitem{qi2018geonet}
Xiaojuan Qi, Renjie Liao, Zhengzhe Liu, Raquel Urtasun, and Jiaya Jia.
\newblock Geonet: Geometric neural network for joint depth and surface normal
  estimation.
\newblock In {\em Proceedings of the IEEE Conference on Computer Vision and
  Pattern Recognition}, pages 283--291, 2018.

\bibitem{qi2020geonet++}
Xiaojuan Qi, Zhengzhe Liu, Renjie Liao, Philip~HS Torr, Raquel Urtasun, and
  Jiaya Jia.
\newblock Geonet++: Iterative geometric neural network with edge-aware
  refinement for joint depth and surface normal estimation.
\newblock {\em arXiv preprint arXiv:2012.06980}, 2020.

\bibitem{qiu2019deeplidar}
Jiaxiong Qiu, Zhaopeng Cui, Yinda Zhang, Xingdi Zhang, Shuaicheng Liu, Bing
  Zeng, and Marc Pollefeys.
\newblock Deeplidar: Deep surface normal guided depth prediction for outdoor
  scene from sparse lidar data and single color image.
\newblock In {\em Proceedings of the IEEE/CVF Conference on Computer Vision and
  Pattern Recognition}, pages 3313--3322, 2019.

\bibitem{ranftl2016dense}
Rene Ranftl, Vibhav Vineet, Qifeng Chen, and Vladlen Koltun.
\newblock Dense monocular depth estimation in complex dynamic scenes.
\newblock In {\em Proceedings of the IEEE conference on computer vision and
  pattern recognition}, pages 4058--4066, 2016.

\bibitem{ren2008local}
Xiaofeng Ren.
\newblock Local grouping for optical flow.
\newblock In {\em 2008 IEEE Conference on Computer Vision and Pattern
  Recognition}, pages 1--8. IEEE, 2008.

\bibitem{revaud2015epicflow}
Jerome Revaud, Philippe Weinzaepfel, Zaid Harchaoui, and Cordelia Schmid.
\newblock Epicflow: Edge-preserving interpolation of correspondences for
  optical flow.
\newblock In {\em Proceedings of the IEEE conference on computer vision and
  pattern recognition}, pages 1164--1172, 2015.

\bibitem{roy2016monocular}
Anirban Roy and Sinisa Todorovic.
\newblock Monocular depth estimation using neural regression forest.
\newblock In {\em Proceedings of the IEEE conference on computer vision and
  pattern recognition}, pages 5506--5514, 2016.

\bibitem{saxena2008make3d}
Ashutosh Saxena, Min Sun, and Andrew~Y Ng.
\newblock Make3d: Learning 3d scene structure from a single still image.
\newblock {\em IEEE transactions on pattern analysis and machine intelligence},
  31(5):824--840, 2008.

\bibitem{song2018edgestereo}
Xiao Song, Xu Zhao, Hanwen Hu, and Liangji Fang.
\newblock Edgestereo: A context integrated residual pyramid network for stereo
  matching.
\newblock In {\em Asian Conference on Computer Vision}, pages 20--35. Springer,
  2018.

\bibitem{wang2020deep}
Jingdong Wang, Ke Sun, Tianheng Cheng, Borui Jiang, Chaorui Deng, Yang Zhao,
  Dong Liu, Yadong Mu, Mingkui Tan, Xinggang Wang, et~al.
\newblock Deep high-resolution representation learning for visual recognition.
\newblock {\em IEEE transactions on pattern analysis and machine intelligence},
  2020.

\bibitem{wang2015towards}
Peng Wang, Xiaohui Shen, Zhe Lin, Scott Cohen, Brian Price, and Alan~L Yuille.
\newblock Towards unified depth and semantic prediction from a single image.
\newblock In {\em Proceedings of the IEEE conference on computer vision and
  pattern recognition}, pages 2800--2809, 2015.

\bibitem{wang2015designing}
Xiaolong Wang, David Fouhey, and Abhinav Gupta.
\newblock Designing deep networks for surface normal estimation.
\newblock In {\em Proceedings of the IEEE Conference on Computer Vision and
  Pattern Recognition}, pages 539--547, 2015.

\bibitem{weickert1998anisotropic}
Joachim Weickert.
\newblock {\em Anisotropic diffusion in image processing}, volume~1.
\newblock Teubner Stuttgart, 1998.

\bibitem{xiao2006bilateral}
Jiangjian Xiao, Hui Cheng, Harpreet Sawhney, Cen Rao, and Michael Isnardi.
\newblock Bilateral filtering-based optical flow estimation with occlusion
  detection.
\newblock In {\em European conference on computer vision}, pages 211--224.
  Springer, 2006.

\bibitem{xu2018pad}
Dan Xu, Wanli Ouyang, Xiaogang Wang, and Nicu Sebe.
\newblock Pad-net: Multi-tasks guided prediction-and-distillation network for
  simultaneous depth estimation and scene parsing.
\newblock In {\em Proceedings of the IEEE Conference on Computer Vision and
  Pattern Recognition}, pages 675--684, 2018.

\bibitem{xu2017multi}
Dan Xu, Elisa Ricci, Wanli Ouyang, Xiaogang Wang, and Nicu Sebe.
\newblock Multi-scale continuous crfs as sequential deep networks for monocular
  depth estimation.
\newblock In {\em Proceedings of the IEEE Conference on Computer Vision and
  Pattern Recognition}, pages 5354--5362, 2017.

\bibitem{xu2019depth}
Yan Xu, Xinge Zhu, Jianping Shi, Guofeng Zhang, Hujun Bao, and Hongsheng Li.
\newblock Depth completion from sparse lidar data with depth-normal
  constraints.
\newblock In {\em Proceedings of the IEEE/CVF International Conference on
  Computer Vision}, pages 2811--2820, 2019.

\bibitem{yang2018unsupervised}
Zhenheng Yang, Peng Wang, Wei Xu, Liang Zhao, and Ramakant Nevatia.
\newblock Unsupervised learning of geometry from videos with edge-aware
  depth-normal consistency.
\newblock In {\em Proceedings of the AAAI Conference on Artificial
  Intelligence}, volume~32, 2018.

\bibitem{yin2019enforcing}
Wei Yin, Yifan Liu, Chunhua Shen, and Youliang Yan.
\newblock Enforcing geometric constraints of virtual normal for depth
  prediction.
\newblock In {\em Proceedings of the IEEE/CVF International Conference on
  Computer Vision}, pages 5684--5693, 2019.

\bibitem{zhang2019pattern}
Zhenyu Zhang, Zhen Cui, Chunyan Xu, Yan Yan, Nicu Sebe, and Jian Yang.
\newblock Pattern-affinitive propagation across depth, surface normal and
  semantic segmentation.
\newblock In {\em Proceedings of the IEEE/CVF Conference on Computer Vision and
  Pattern Recognition}, pages 4106--4115, 2019.

\end{thebibliography}
